\documentclass[conference]{IEEEtran}
\IEEEoverridecommandlockouts
\usepackage{cite}
\usepackage{amsmath,amssymb,amsfonts}
\usepackage{algorithmic}
\usepackage{graphicx}
\usepackage{textcomp}
\usepackage{enumitem}
\usepackage{color}
\usepackage{url}
\usepackage{multirow}

\usepackage[linesnumbered]{algorithm2e}
\usepackage{graphicx}
\usepackage{textcomp}
\usepackage{enumitem}
\usepackage{url}
\usepackage{floatrow}
\usepackage{wrapfig}
\usepackage{algorithmic}
\usepackage{multirow}

\usepackage{array}
\newcolumntype{M}[1]{>{\centering\arraybackslash}m{#1}}

\def\BibTeX{{\rm B\kern-.05em{\sc i\kern-.025em b}\kern-.08em
    T\kern-.1667em\lower.7ex\hbox{E}\kern-.125emX}}

\newcommand{\Comment}[1]{}
\newcommand{\sig}[1]{\textsf{{#1}}}

\newtheorem{Definition}{Definition}

\begin{document}

%\sisetup{quotient-mode=fraction}

\title{Runtime Monitoring Neuron Activation Patterns
}

\author{\IEEEauthorblockN{Chih-Hong Cheng\IEEEauthorrefmark{1}, Georg N\"{u}hrenberg\IEEEauthorrefmark{1} and Hirotoshi Yasuoka\IEEEauthorrefmark{2}}
	\IEEEauthorblockA{\IEEEauthorrefmark{1}fortiss - Research Institute of the Free State of Bavaria 
		}
	\IEEEauthorblockA{\IEEEauthorrefmark{2}DENSO CORPORATION\\
		Email: \texttt{\{cheng,nuehrenberg\}@fortiss.org, hirotoshi\_yasuoka@denso.co.jp}
	}
}

\maketitle

\vspace{-5mm}

\begin{abstract}
	
For using neural networks in safety critical domains, it is important to know if a decision made by a neural network is supported by prior similarities in training. We propose runtime neuron activation pattern monitoring - after the standard training process, one creates a monitor by feeding the training data to the network again in order to store the neuron activation patterns in abstract form. In operation, a classification decision over an input is further supplemented by examining if a pattern similar (measured by Hamming distance) to the generated pattern is contained in the monitor.
If the monitor does not contain any pattern similar to the generated pattern, it raises a warning that the decision is not based on the training data.
Our experiments show that, by adjusting the similarity-threshold for activation patterns, the monitors can report a significant portion of misclassfications to be not supported by training with a small false-positive rate, when evaluated on a test set.

\end{abstract}


\section{Introduction}~\label{sec.introduction}

\vspace{-5mm}
For highly automated driving, neural networks are the \emph{de facto} option for vision-based perception. Nevertheless, one fundamental challenge for using neural networks in such a safety-critical application is to understand if a trained neural network performs inference 
``\emph{outside its comfort zone}''. This appears when the network 
needs to significantly extrapolate from what it learns (or remembers) from the training data, as similar data has not appeared in the training process. 

\begin{figure}[t]
	\centering
	\includegraphics[width=\columnwidth, trim=5.2cm 2cm 3.8cm 1.3cm, clip]{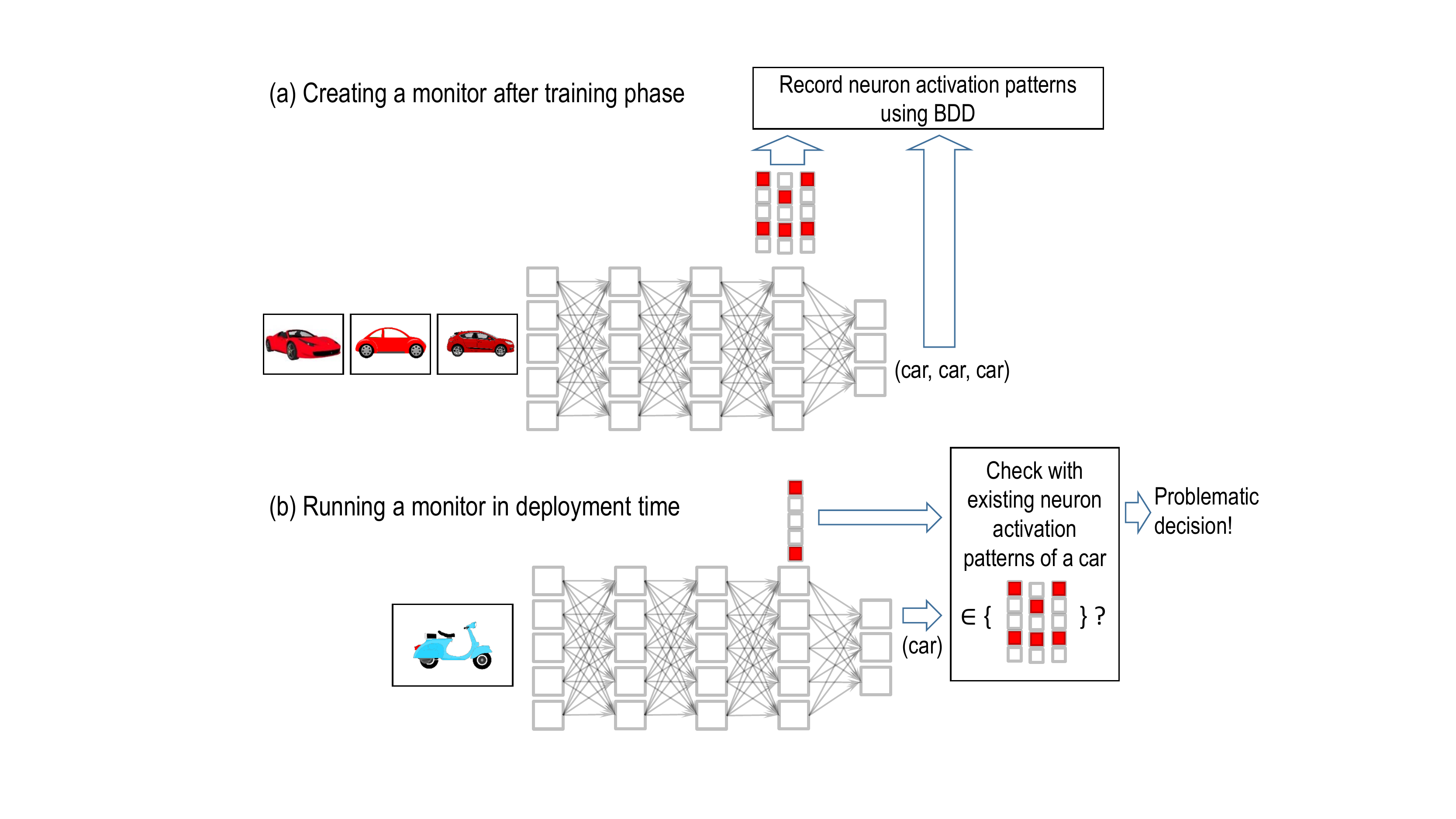}
	\vspace{-3mm}
	\caption{High-level workflow on runtime monitoring neuron activation patterns.}
	\label{fig.monitoring}
\end{figure}

In this paper, we address this problem by \emph{runtime monitoring neuron activation patterns},  where the underlying workflow is illustrated in Figure~\ref{fig.monitoring}.
After completing the training process, one records the neuron activation patterns for close-to-output neural network layers for all correctly predicted data used in the training process. Neurons in close-to-output layers in general represent high-level features, as demonstrated by recent approaches in interpreting neural networks~\cite{zeiler2014visualizing}. 
As state-of-the-art neural networks commonly use ReLU or its variations as activation function, we select the ReLU \emph{on-off activation pattern} to record the presence or absence of high-level features. At the same time, on-off patterns allow efficient storage using binary decision diagrams (BDDs)~\cite{bryant1992symbolic}. 
In operation, a classification decision is supplemented with a BDD-based monitor to detect whether the provided input has triggered an unseen neuron activation pattern - whenever an unseen activation pattern appears, the decision made by the neural network is considered to be less reliable.
For the example in Figure~\ref{fig.monitoring}-(b), the scooter is classified as a car, but as its neuron activation pattern is not among the existing patterns created from the training data, the monitor reports that the decision made by the neural network can be problematic.
The frequent appearance of unseen patterns provides an indicator of data distribution shift to the development team; such information is helpful as it may indicate that a neural network deployed on an autonomous vehicle needs to be updated.

% trim={<left> <lower> <right> <upper>}

\begin{figure}[t]
	\centering
	\includegraphics[width=0.7\columnwidth]{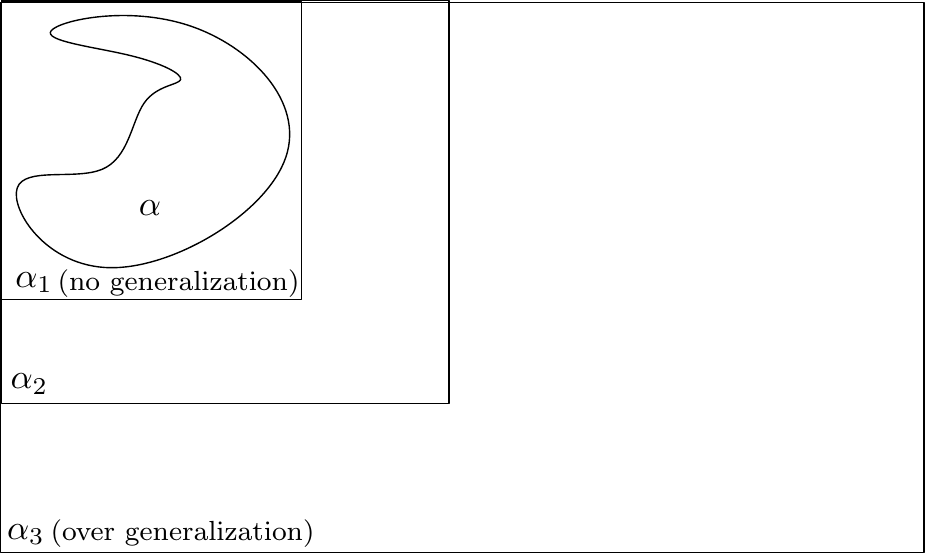}
	\caption{Finding ``just-right'' abstraction for runtime monitors}
	\label{fig.abstraction}
\end{figure}

Nevertheless, for such an approach to be useful, we  encounter technical difficulties where in the created monitor,  the coarseness of abstraction should be \emph{abstract enough, but not too abstract}.  An illustration can be found in Figure~\ref{fig.abstraction}, where given~$\alpha$ to be all visited states from the training data, an abstraction such as~$\alpha_1$ allows nearly no generalization effect, making all encountered data in operation time to be ``not visited''. On the other hand, an abstraction such as~$\alpha_3$ is too coarse in that every observed pattern in operation time is identified to be ``visited''; such a monitor is also not useful. Overall, we have applied the following techniques to control the coarseness of abstraction.

\begin{description}
	\item[(Enlarge the abstraction)] Apart from merely including visited patterns, we further develop technologies to enlarge the pattern space by considering all neuron activation patterns whose Hamming distance with existing patterns are within a certain threshold. It can also be efficiently implemented using existential quantification as commonly seen in many BDD software packages. Adding additional patterns does not influence performance - the membership query during runtime remains in the worst case in time linear to the number of neurons in the monitored layer (due to the use of BDDs). In addition, we apply gradient-based sensitivity analysis~\cite{simonyan2013deep} to only monitor important neurons, thereby allowing unmonitored neurons to hold arbitrary values in the abstraction. This also overcomes the limitation where the maximum number of BDD variables one can use in practice is around hundreds.
	
	\vspace{1mm}
	\item[(Infer when to stop enlarging)] To ensure that the abstraction is not too coarse, we take a validation set (which is expected to have the same distribution as in operation, but with ground-truth labels) and gradually increase the Hamming distance such that in the created region of abstraction, whenever the occurrence of out-of-pattern scenarios appears, it is also likely that misclassification appears. We applied this concept to decide the coarseness of abstraction for classifying standard image benchmarks such as MNIST~\cite{lecun1998mnist} German Traffic Sign Recognition Benchmark (GTSRB)~\cite{stallkamp2011german}, as well as a vision-based front-car detector for automated highway piloting.
\end{description}

\vspace{1mm}
The rest of the paper is structured as follows. Section~\ref{sec.building.monitor} describes how to build neuron activation pattern monitors with the use of BDDs. Section~\ref{sec.control.abstraction} gives examples in terms of controlling the coarseness of abstraction.  
We summarize related work in Section~\ref{sec.related} and conclude in Section~\ref{sec.conclusion} with further research directions.

\section{Building Neuron Activation Pattern Monitors}\label{sec.building.monitor}

We describe the underlying principles of our runtime monitoring approach for neural networks. For simplicity, the presented algorithm is for image classification, and we focus on runtime monitoring fully-connected neural network layers.
Monitoring convolutional layers can be achieved by treating layers having convolutional filters as layers with fully connected neurons where missing connections are assigned with zero weights.

A neural network is comprised of $L$ layers
where operationally,  the $l$-th layer for $l\in\{1,\dots,L\}$ of the network is a function $g^{(l)}: \mathbb{R}^{d_{l-1}} \rightarrow \mathbb{R}^{d_{l}}$, with $d_{l}$ being the dimension of layer~$l$.  
Given an input $\sig{in} \in \mathbb{R}^{d_{0}}$, the output of the $l$-th layer of the neural network $f^{(l)}$ is given by the functional composition of the $l$-th layer and previous layers $f^{(l)}(\sig{in}) := \circ_{i=1}^{(l)} g^{(i)}(\sig{in})  = g^{(l)}(\ldots g^{(2)}(g^{(1)}(\sig{in})))$. 
For a neural network classifying $C$ categories $d_{L} = C$. Given the computed output $f^{(L)}(\sig{in}) = (v_1, \ldots, v_C)$,  the \emph{decision} $\sig{dec}_{f^{(L)}}(\sig{in})$ of classifying input~$\sig{in}$ to a certain class is based on choosing the index~$i$ with the maximum value $v_i$ among elements in the output vector, i.e., $\sig{dec}_{f^{(L)}}(\sig{in}):= \sig{argmax}_{i\in \{1, \ldots, C\}} \{v_1, \ldots, v_C\}$.

An important case in modern neural networks is the use of layers implementing \emph{Rectified Linear Unit} (ReLU), where the corresponding function~$g^{(l)}$ maintains the input dimension and transforms an input vector element-wise by keeping its positive part, i.e., $g^{(l)}(v_1, \ldots, v_{d_{l-1}}) := (v_1', \ldots, v_{d_{l-1}}')$ where  $v_i' := \sig{max}(0, v_i)$ for $i\in \{1, \ldots, d_{l-1}\}$.

By interpreting an input element $v_i$ to the ReLU layer as  feature intensity, if  $v_i$ has value greater than zero, then it is considered to be \emph{activated}, while  $v_i$ having value less or equal to zero is considered to be \emph{suppressed} by ReLU. 
With this intuition in mind, our definition of a \emph{neuron activation pattern} is based on capturing the activation and suppression of features. 

\begin{Definition}[\textbf{Neuron activation pattern}] Given a neural network with input~$\sig{in}$ and the $l$-th layer being ReLU,
	$\sig{pat}(f^{(l)}(\sig{in}))$, the neuron activation pattern at layer $l$, is defined as follows:
	\begin{equation*}
	\sig{pat}(f^{(l)}(\sig{in})) :=  (p_{\mathrm{relu}}(v_1),\dots ,p_{\mathrm{relu}}(v_{d_{l}}))
	\end{equation*}
	where $(v_1,\dots ,v_{d_{l}}) = f^{(l)}(\sig{in})$ is the output from layer $l$, and $p_{\mathrm{relu}}:\mathbb{R}\to \{\sig{0},\sig{1}\}$ captures the activation cases:
	
	\vspace{-3mm}
	\begin{equation*}
	p_{\mathrm{relu}}(x)=\begin{cases}\sig{1} & x > 0 \\ \sig{0} & otherwise \end{cases}    
	\end{equation*}
	
\end{Definition}

Let $\mathcal{T}$ denote the set of training inputs and let $\mathcal{T}_{c} \subseteq \mathcal{T}$ denote the set of all training images labelled as class~$c$ based on the ground truth. For each class~$c$, we define the corresponding ``comfort zone'' for a neural network to be the set of activation patterns visited for all correctly classified training images, together with other neuron activation patterns that are close (via Hamming distance) to visited patterns.

\begin{Definition}[\textbf{$\gamma$-comfort zone}]
	Given a neural network and its training set $\mathcal{T}$, the $\gamma$-comfort zone $\mathcal{Z}^{\gamma}_{c}\subseteq \{\sig{0},\sig{1}\}^{d_{l}}$ for classifying class~$c$,  under the condition where the $l$-th layer is ReLU, is defined recursively as follows:  
	
\vspace{-2mm}
	\[
\mathcal{Z}^{\gamma}_{c} := 
\begin{cases}
\{\sig{pat}(f^{(l)}(\sig{in}))\ |\ \sig{in}\in \mathcal{T}_{c} \wedge \sig{dec}_{f^{(L)}}(\sig{in})=c\}, \text{if } \gamma = 0\\
\mathcal{Z}^{\gamma -1}_{c} \cup  \{\sig{p}\,|\,
\sig{p} \in \{\sig{0},\sig{1}\}^{d_{l}} \wedge \\\;\;\;\; \;\;\;\; \;\;\;\; \;\;\;\; \;\;\;\; \;\;\;
\exists \sig{p'} \in \mathcal{Z}^{\gamma -1}_{c}:\mathcal{H}(\sig{p},\sig{p'})=1 \},    \text{if } \gamma > 0
\end{cases}
\]

	\noindent where $\mathcal{H}(p, p')$ is the function to compute the Hamming distance between two pattern vectors $\sig{p}, \sig{p}' \in \{0,1\}^{d_{l}}$.
	
\end{Definition}

Lastly, a neuron activation pattern monitor stores the computed comfort zone for each class using the training data. 

\begin{Definition}[\textbf{Neural activation pattern monitor}]
	Given a neural network for classifying $C$ classes,  its training set and a user-specified $\gamma$, its neuron activation pattern monitor is defined as
	$\langle \mathcal{Z}^{\gamma}_{1}, \ldots, \mathcal{Z}^{\gamma}_{C}\rangle$. 
\end{Definition}

Note that as $\mathcal{Z}^{\gamma}_c\subseteq \{\sig{0},\sig{1}\}^{d_{l}}$, the construction of $\mathcal{Z}^{\gamma}_c$ can be done using binary decision diagrams with $d_{l}$ variables. Algorithm~\ref{algo:build_monitor} describes how to construct such a monitor, where \sig{bdd.emptySet}, \sig{bdd.or}, and \sig{bdd.encode} are functions used to create an empty set, to perform set union, and to encode an activation pattern into BDDs. The function $\sig{bdd.exists}(j, \sig{set})$  performs the existential quantification on $\sig{set}$ over the $j$-th variable.

In Algorithm~\ref{algo:build_monitor}, lines~4 to~8 record all visited patterns to form $\mathcal{Z}^{0}_c$. Subsequently, lines~9 to~14 build $\mathcal{Z}^{i}_c$ from $\mathcal{Z}^{i-1}_c$. In particular, computing the enlarged~$\mathcal{Z}^{\gamma}_c$ from $\mathcal{Z}^{\gamma -1}_c$ can be efficiently achieved using the existential quantification operation as listed in line~12. Consider an example where $\mathcal{Z}^0_c = \{\sig{001}\}$, then  the operation $\sig{bdd.exists}(j, \mathcal{Z}^{0}_{c})$, for $j=1,2,3$, creates $\{\sig{-01}\}, \{\sig{0-1}\}, \{\sig{00-}\}$ respectively. The union over existentially quantified result creates an enlarged set containing additional patterns with Hamming distance equal to~$1$.

\begin{algorithm}[t]
	\KwIn{neural network and $l$-th layer  to monitor, training set $\mathcal{T}$, user specified $\gamma$}
	\KwOut{runtime activation pattern monitor $\langle \mathcal{Z}^{\gamma}_{1}, \ldots, \mathcal{Z}^{\gamma}_{C}\rangle$}
	\tcc{initialize monitors as empty BDDs}
	\For{$c\in C$}{$\mathcal{Z}^0_c \gets \sig{bdd.emptySet()}$}
	\tcc{iterate all images}
	\For{$\sig{in}\in \mathcal{T}$}{
		%\tcc{iterate all images of a class}
		
		\tcc{check if prediction is correct}
		\If{$\sig{dec}_{f^{(L)}}(\sig{in})=c \wedge \sig{in} \in \mathcal{T}_c$}{
			\tcc{add activation pattern to the corresponding BDD}
			$\mathcal{Z}^0_c \gets \sig{bdd.or}(\mathcal{Z}^0_c,   \sig{bdd.encode}(\sig{pat}(f^{(l)}(\sig{in}))))$
	}}
	\lFor{$c = 1, \ldots, C$, $i = 1, \ldots, \gamma$}{
		$\mathcal{Z}^{i}_c \gets \sig{bdd.emptySet()}$}
	\For{$c = 1, \ldots, C$}{
		\For{$i = 1, \ldots, \gamma$}{
			\lFor{$j = 1, \ldots, d_l$}{
				$\mathcal{Z}^i_c \gets \sig{bdd.or}(\mathcal{Z}^i_c,   \sig{bdd.exists}(j, \mathcal{Z}^{i-1}_{c}))$
			}    
		}
	}
	
	\Return{$\langle \mathcal{Z}^{\gamma}_{1}, \ldots, \mathcal{Z}^{\gamma}_{C}\rangle$}
	\vspace{2mm}
	\caption{Building a neuron activation pattern monitor after training}
	\label{algo:build_monitor}
\end{algorithm}

\vspace{2mm}
\noindent \textbf{(Neuron selection via gradient analysis)} For layers with large neuron amounts, as the use of BDD has practical variable limits around~$200$, one extension is to only monitor the activation patterns over a  subset of neurons that are important for the classification decision. 
One way of selecting neurons to be monitored is to apply   gradient-based sensitivity analysis similar to the work of saliency map~\cite{simonyan2013deep}. The underlying principle is that for the output of neuron $n_i$ over neuron $n_c$ producing output class~$c$, one computes $\frac{\partial n_c}{\partial n_i}$. Subsequently, one only selects  neuron $n_i$ if $|\frac{\partial n_c}{\partial n_i}|$ is large, as the change of  value~$n_i$ significantly influences the output~$c$ due to the  derivative term. 

As a special case, if one monitors patterns over the neuron layer immediately before the output layer, and there is no non-linear activation in the output layer (which is commonly seen in practice), $\frac{\partial n_c}{\partial n_i}$ is simply the weight connecting~$n_i$ to~$n_c$.

\vspace{-2mm}
\section{Controlling the Abstraction}\label{sec.control.abstraction}
\vspace{-2mm}

As stated in the introduction, the coarseness of abstraction should be carefully designed to make the resulting monitor useful. Both the number of neurons being monitored and the value~$\gamma$ are hyper-parameters to control the coarseness of abstraction. We have implemented the concept to examine the effect of different~$\gamma$ using the PyTorch machine learning framework\footnote{Pytorch: \url{https://pytorch.org/}} and the python-based BDD package \texttt{dd}\footnote{\texttt{dd}: \url{https://pypi.org/project/dd/}}.

Based on two publicly available image classification datasets MNIST~\cite{lecun1998mnist} and GTSRB~\cite{stallkamp2011german}, we trained two neural networks. The architectures of the networks are summarized in Table~\ref{table.networks.trained}. After training, we build the runtime monitors based on Algorithm~\ref{algo:build_monitor}. For network~2, in the experiment we (i) only construct the monitor for the stop sign ($c=14$) and (ii) out of $84$ neurons in a layer, only~$25\%$ are monitored based on gradient-based analysis. We have gradually increased $\gamma$ and recorded the rate of out-of-pattern images for all validation images, as well as the portion of misclassified images within out-of-pattern images.

\begin{table}[t]
	\centering
		\begin{small}
	\begin{tabular}{|c|c|p{4.3cm}|p{2.1cm}|}

		\hline
		ID  & Classifier & Model architecture & Accuracy (train/validation) \\
		\hline
		1  & MNIST & ReLU(Conv(40)), MaxPool, \newline ReLU(Conv(20)), MaxPool, \newline ReLU(fc(320)), ReLU(fc(160)), \newline ReLU(fc(80)),  \textbf{ReLU(fc(40))}, \newline fc(10) & $99.34\%$, $98.81\%$ \\\hline
		2 & GTSRB & ReLU(BN(Conv(40))), MaxPool, \newline ReLU(BN(Conv(20))), MaxPool, \newline ReLU(fc(240)), \textbf{ReLU(fc(84))},\newline fc(43) & $99.98\%$, $96.73\%$ \\[0.2cm]\hline	

	\end{tabular}
		\end{small}
	\vspace{-3mm}
	\caption{Architectures and accuracies of the networks used in the experiment.
	Convolutional layers (Conv) have kernel size $(5, 5)$ and stride $(1, 1)$. We use $2\times2$ max pooling layers (MaxPool).
	Fully-connected layers and batch normalization are denoted by $\mathrm{fc}(\cdot)$ and $\mathrm{BN}(\cdot)$.
	The layer being monitored is highlighted in bold text.
}
\label{table.networks.trained}
\end{table}

\begin{table}[t]
	\centering
	\begin{small}
	\begin{tabular}{|M{0.3cm}|M{1.7cm}|M{0.2cm}|M{2.1cm}|M{3.2cm}|}
		\hline
		ID & {\scriptsize misclassification rate} & $\gamma$ & $\frac{\#\emph{out-of-pattern images}}{\#\emph{total images}}$  &  $\frac{\#\emph{out-of-pattern misclassified images}}{\#\emph{out-of-pattern images}}$ \\
		
		\hline
		\multirow{3}{*}{1}  & \multirow{3}{*}{$1.19\%$} & $0$ & $7.66\%$ & $10.70\%$ \\
		& & $1$  & $2.01\%$ &$21.89\%$ \\
		& & $2$  & $0.6\%$ &$31.66\%$ \\
		\hline
		\multirow{4}{*}{2}  & \multirow{4}{*}{$3.27\%$} & $0$  & $32.92\%$ & $10.13\%$ \\
		& & $1$  & $15.0\%$ & $19.44\%$  \\ 
		& & $2$  & $7.08\%$ & $41.17\%$ \\ 
		& & $3$  & $4.58\%$ & $54.54\%$ \\ 
		\hline
	\end{tabular}
	\end{small}
\vspace{-3mm}
	\caption{Results of applying runtime neuron activation monitoring}
	\label{tab:accuray_results}
\end{table}

For network~1 classifying MNIST, the rates of $\frac{\#\emph{out-of-pattern images}}{\#\emph{total images}}$ for all $\gamma \in \{0,1,2\}$ are all relatively small. For network~2 classifying GTSRB, one can argue that the abstraction using $\gamma = 0$ is not coarse enough, as the network has a low mis-classification rate (around $3.27\%$) but the monitor reports that around $32.92\%$ of the images create patterns that are not included in the monitor.

\begin{description}
	\item[(MNIST with $\gamma=2$)] If there is no distributional shift in operation, the monitor will not signal problems in $99.4\%$ ($100\% - 0.6\%$) of its overall operation time, implying that it is largely silent. Nevertheless, whenever it signals an issue of unseen patterns, apart from arguing that the network is making a decision without prior similarities, one may even argue that there is a non-neglectable probability of $31.66\%$ where the decision being made by the network is problematic\footnote{The argument is based on an assumption where no distributional shift implies that $\frac{\#\emph{out-of-pattern misclassified images}}{\#\emph{out-of-pattern images}}$ remains the same in validation and in operation.}, although the neural network may still report that the input is classified to the class with a high probability.

	\item[(GTSRB with $\gamma=3$)]  If there is no distributional shift in operation, the monitor will not signal problem in $95.42\%$ ($100\% - 4.58\%$) of its operating time. Whenever it signals  an issue of unseen patterns, there is a non-neglectable probability of $54.54\%$ where it is indeed misclassified. 
\end{description}

\begin{figure}[t]
	\centering
	\includegraphics[width=\columnwidth, trim=0cm 0.5cm 0cm 0cm, clip]{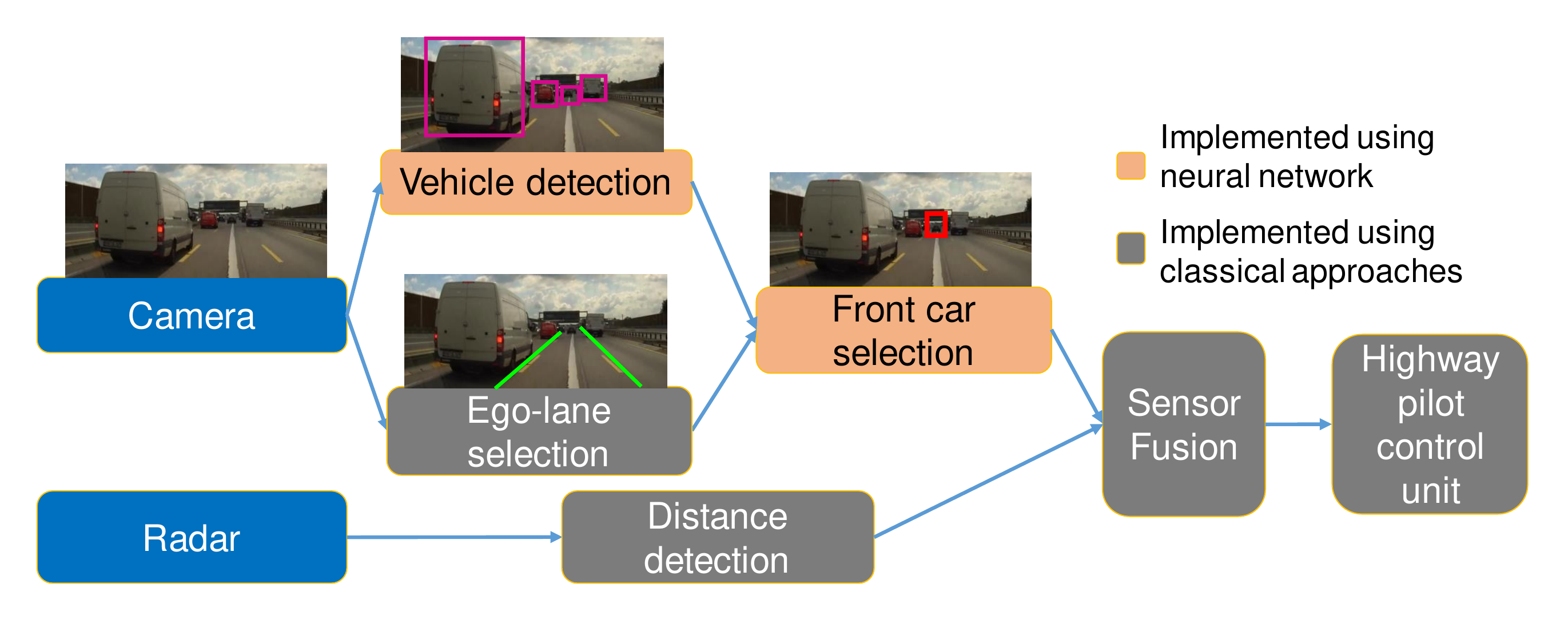}
	\caption{High-level architecture of a front-car detection unit for a highway piloting system.}
	\label{fig.pilot}
\end{figure}

\vspace{1mm}
\noindent \textbf{(Case Study)} We also experimented the runtime monitoring technique on a  vision-based front-car detection system for highway piloting. The vision subsystem (cf. Figure~\ref{fig.pilot}) contains three components: (1) vehicle detection, (2) lane detection, and (3) front-car selection. The front-car selection unit is implemented using a neural network-based classifier, which takes the lane information and the bounding box of vehicles, and produces either an index of the bounding vehicle or a special class ``$\sharp$" for which no forward vehicle is considered to be a front car.  

% trim={<left> <lower> <right> <upper>}

\vspace{-2mm}
\section{Related Work}\label{sec.related}

Using neural networks in safety critical applications has raised needs for creating dependability claims. Recent results in compile-time formal verification techniques such as RuLUplex~\cite{katz2017reluplex} or Planet~\cite{ehlers2017formal} use constraint solving  to examine if for all inputs within a bounded polyhedron, it is possible for the network to generate undersired outputs. These techniques are used when a risk property is provided by domain experts beforehand, and they are only limited to piecewise linear networks with a few number of neurons. Our work of neuron monitoring is more related to the concept of runtime verification~\cite{leucker2009brief}, which examines if a runtime trace has violated a given property. The  generalizability condition, as defined by the $\gamma$-comfort zone created after training, can be understood as a safety property. To the best of our knowledge, we are unaware of any work in runtime verification that considers the problem of generalizability monitoring of neural networks. In terms of scalability, our framework also allows taking arbitrary large networks with other nonlinear activation functions, so long as the neurons being monitored are ReLU.

Lastly, within machine learning (ML), the work of filtering adversarial attacks~\cite{grosse2017statistical,xu2017feature} reply on creating another  ML component to perform detection (thus preventing the network from making wrong decisions). Our proposed method differs from these ML-based approach in that the \emph{sound} over-approximation of the visited inputs implies that if the monitor reports the occurrence of an unseen pattern, the occurrence is always genuine. The \emph{sure guarantee} (in contrast to concepts such as \emph{almost-sure}\footnote{\url{https://en.wikipedia.org/wiki/Almost_surely}} which is the best one can derive with statistical machine learning methods) makes the certification of such a monitor in the safety domain relatively easier. In particular within the domain of autonomous driving, it is highly likely that the test set used in engineering time will deviate from the real world data (the black swan effect), making any probabilistic claim hard to be certified. 

\vspace{-1mm}
\section{Concluding Remarks}\label{sec.conclusion}

In this paper, we proposed neuron activation pattern monitoring as a method to detect if a decision made by a neural network is not supported by  prior similarities in training. 
We envision that a neuron activation pattern monitor can be served as a medium to assist the sensor fusion process on the architecture level, as a decision made by the network may not be fully trusted due to no ground-truth being offered in operation time. 

The established connection between formal methods and machine learning also reveals several possible extension schemes. (1) The technique shall be directly applicable on object detection networks such as YOLO~\cite{redmon2016you}, whose underlying principle is to partition an image to a finite grid, with each cell in the grid offering object proposals.
(2) We are also studying the feasibility on more refined domains using tools such as difference bound matrices~\cite{mine2001new}, in order to better capture an abstract representation of the visited activation patterns.

\Comment{

\begin{itemize}
	\item Currently our experiment is restricted to classification, but the technique shall be directly applicable on object detection networks such as YOLO~\cite{redmon2016you}, whose underlying principle is to partition an image to a finite grid, with each cell in the grid offering object proposals.
	\item We are also studying the feasibility on more refined domains using tools such as difference bound matrices~\cite{mine2001new}, in order to better capture an abstract representation of the visited activation patterns.
\end{itemize}

}

%\bibliographystyle{abbrv}
%\bibliography{references} 

\end{document}